\documentclass{article}

\usepackage[preprint]{styles/neurips_2024}

\usepackage[utf8]{inputenc} 
\usepackage[T1]{fontenc}    
\usepackage{hyperref}       
\usepackage{url}            
\usepackage{booktabs}       
\usepackage{amsfonts}       
\usepackage{nicefrac}       
\usepackage{microtype}      
\usepackage{xcolor}         
\usepackage{amsmath}
\usepackage{graphicx}
\usepackage{adjustbox}

\title{Entity-based Reinforcement Learning\\ for Autonomous Cyber Defence}

\author{%
  Isaac Symes Thompson \\
  The Alan Turing Institute\\
  London\\
  \texttt{isymesthompson@turing.ac.uk} \\
    \And
  Alberto Caron \\
  The Alan Turing Institute\\
  London\\
  \texttt{acaron@turing.ac.uk} \\
  \And
  Vasilios Mavroudis \\
  The Alan Turing Institute\\
  London\\
  \texttt{vmavroudis@turing.ac.uk} \\
  \And
  Chris Hicks \\
  The Alan Turing Institute\\
  London\\
  \texttt{chicks@turing.ac.uk} \\
}

\begin{document}

\maketitle

\begin{abstract}
    A significant challenge for autonomous cyber defence is ensuring a defensive agent's ability to generalise across diverse network topologies and configurations. This capability is necessary for agents to remain effective when deployed in dynamically changing environments, such as an enterprise network where devices may frequently join and leave. Standard approaches to deep reinforcement learning, where policies are parameterised using a fixed-input multi-layer perceptron (MLP) expect fixed-size observation and action spaces. In autonomous cyber defence, this makes it hard to develop agents that generalise to environments with network topologies different from those trained on, as the number of nodes affects the natural size of the observation and action spaces. To overcome this limitation, we reframe the problem of autonomous network defence using entity-based reinforcement learning, where the observation and action space of an agent are decomposed into a collection of discrete entities. This framework enables the use of policy parameterisations specialised in compositional generalisation. We train a Transformer-based policy on the Yawning Titan cyber-security simulation environment and test its generalisation capabilities across various network topologies. We demonstrate that this approach significantly outperforms an MLP-based policy when training across fixed-size networks of varying topologies, and matches performance when training on a single network. We also demonstrate the potential for zero-shot generalisation to networks of a different size to those seen in training. These findings highlight the potential for entity-based reinforcement learning to advance the field of autonomous cyber defence by providing more generalisable policies capable of handling variations in real-world network environments.
\end{abstract}

\section{Introduction}

The development of autonomous agents for cyber defence is an area of research that aims to address the increasing complexity and volume of cyber threats. Solutions based on deep reinforcement learning (RL), inspired by high-profile successes in other complex domains~\citep{AlphaStarNature, Roy_2021, bello2017neuralcombinatorialoptimizationreinforcement}, offer the potential to rapidly detect, analyse, and respond to cyber attacks at scale~\citep{nguyen2021deep}. A fundamental challenge involved in the deep RL approach is to develop agents that are generalisable to variable network topologies. This is difficult to achieve with fixed-input neural network architectures commonly used for function approximation and standard RL interfaces such as Gym or Gymnasium \citep{OpenAIGym,towers2024gymnasium}, which expect fixed-size observation vectors and action spaces. The lack of this generalisability imposes serious limitations on the potential deployment of RL agents in real-world network scenarios, as enterprise networks can vary dramatically in size, structure, and complexity.

One way of approaching RL for autonomous cyber defence is to construct an observation vector by concatenating features from all of the nodes in the network, and applying standard deep RL algorithms with policies or value functions parameterised using multi-layer perceptrons (MLP). Such an agent can be trained in environments with fixed-size observation and action spaces~\citep{cage_challenge_2}, but would not be readily transferable to other configurations. For example, under this approach, it would be impossible to deploy an agent on networks larger than the ones encountered during training, 
and any attempts to adapt to different networks that are the same size or smaller using padding are unlikely to be successful due to the `non-exchangeability' of input features ~\citep{mern_object_exchangeability_paper,mern_object_exchangeability_abstract}. That is, the nature of nodes and connections between them also affect the meaning of elements of the agent's observations, with elements in the same position in the input vector to the policy network not necessarily representing a single consistent object. Also, actions indexed in a discrete action space may not have a consistent effect depending on the objects they interact with. This places a large burden on an MLP to learn these interactions from scratch, which may require an impractically large number of environment samples.

This problem falls into the broad category of \textit{compositional generalisation}, which is a long-studied problem in reinforcement learning \citep{BOUTILIER_factoredMDPs, Guestrin_2003, Diuk_Object_Oriented_MDP_paper}. Recently, a number of works have explored introducing compositional generalisation to deep reinforcement learning ~\citep{compositional_generalization_survey_baihan_lin_2023,mambelli2022compositionalmultiobjectreinforcementlearning, zhou2022policyarchitecturescompositionalgeneralization, haramati2024entitycentricreinforcementlearningobject}, through using permutation invariant policy architectures based on attention~\citep{Transformers} or Deep Sets~\citep{zaheer2017deepsets}. Indeed, some existing work in autonomous cyber defence has suggested using Transformers to parameterise a value function \citep{mern2021reinforcementlearningindustrialcontrol_arxiv}, or using graph neural networks for policies \citep{janisch2023nasimemu, janisch2023symbolicrelationaldeepreinforcement, palmer2023deep} which are invariant to the number of nodes in the environment.

This work aims to address this challenge by framing autonomous cyber defence as an entity-based reinforcement learning problem~\citep{winter2023entity}. Entity-based RL is a paradigm where environments are decomposed into collections of discrete entities which an agent observes and acts upon directly. In the context of a network environment, each node of the network can be treated as an entity, with a defending agent's observation space permitted to vary between environment instances according to how many nodes are visible. With a suitable training regime, it should be possible for an agent trained using entity-based RL to generalise to any network topology, provided the network in question contains entities of familiar types. The Entity Gym framework \citep{entity-gym}, intended to occupy a similar role as OpenAI Gym or Gymnasium, provides a standard interface in which to cast entity-based environments, and allows any agent designed for entity-based reinforcement learning to be seamlessly deployed in such environments. 

In this paper, we introduce a wrapper for the Yawning Titan cyber-simulation environment~\citep{Yawning_Titan} that adheres to the Entity Gym interface.
In this environment, we trained and compared two types of policies: a Transformer-based  policy, RogueNet~\citep{winter2023entity}, trained using the Entity Gym interface and a multilayer perceptron (MLP) policy trained on the standard Gym~\citep{OpenAIGym} interface for Yawning Titan. We trained both types of policies using Proximal Policy Optimization (PPO) \citep{schulman2017proximal} with two different training modes, which we refer to as \textit{Random} and \textit{Static}, across network sizes of 10, 20, and 40 nodes. In the \textit{Random} mode, a new random network topology is generated with each environment reset, whilst in the \textit{Static} mode, a random network topology is initialised at the start of training and kept constant throughout. This approach has been designed to facilitate comparison with the MLP-based policy, which is able to train well in the \textit{Static} mode, but struggles with \textit{Random} training where it encounters a new network configuration with every episode. We also performed zero-shot evaluation of the Transformer-based policies on network sizes not seen during training, something not possible with standard MLP-based policies. 
From these experiments, we find that in this environment, entity-based policies outperform MLP-based policies during training, particularly in the \textit{Random} mode where the MLP-based policy struggled to learn. We also find that entity-based policies generalise well to networks of unseen sizes, performing similarly to policies trained natively in those environments.

The paper is organised as follows. In Section 2 we give an overview of motivating background and related literature. In Section 3 we describe the Entity Gym and the RogueNet architecture \citep{winter2023entity, entity-gym}, which we have used to construct our environment and train entity-based policies. In Section 4 we provide a description of the Yawning Titan simulator and choices we have made for the environment setting for our experiments. In Section 5 we explain how we have adapted the Yawning Titan simulator to the entity-based framework. In Section 6 we outline the methodology and experiments we have chosen to run. In Section 7, we present the results of our experiments and conclude with a discussion of the results and future work in Section 8.

The code for our Entity Gym adaptation of Yawning Titan, as well as the scripts used to run the experiments in this paper, are available in the following repository: \url{https://github.com/alan-turing-institute/Entity-Based-Yawning-Titan}.

\section{Background and related work}

\subsection{Autonomous Cyber Defence}
Over the last decade, a number of simulators have been developed to facilitate research into using reinforcement learning for autonomous cyber defence \citep{cage_challenge_1, cage_challenge_2, cage_challenge_4_announcement, standen2021cyborg, FARLAND, Yawning_Titan, msft:cyberbattlesim}. Typically, these involve a network scenario consisting of nodes of varying types. A defensive \textbf{blue} agent is provided with a fixed observation and action space with which to learn to defend the network from an attacking \textbf{red} agent, whose behaviour is unintelligent and part of the environment dynamics.

The CybORG environment~\citep{cage_cyborg_2022} was initially conceived as part of a competition, Cage Challenge 1~\citep{cage_challenge_1}, intended to encourage research into RL-based autonomous cyber defence agentsngExamples of h solutions to the challenge include using agents trained with PPO \citep{schulman2017proximal}, combined with a bandit controller to select between policies specialised in specific attack patterns~\citep{foley2022autonomous,canaries2023,foley2023inroadsautonomousnetworkdefence,bates2023reward}, or with an ensemble approach \citep{WOLK_beyond-cage_generalisation}.
Whilst not the focus of our paper, the CybORG environment features several types of nodes that could be used as entity types in the entity-based RL setting; \textit{User Host}, \textit{Enterprise Server}, \textit{Operational Host} and the \textit{Operational Server}.

Recognising that training and evaluating on a fixed network environment is not satisfactory for real-world applicability, a number of works have explored forms of generalisation. For example, to unseen red agent dynamics \citep{WOLK_beyond-cage_generalisation}, or to unseen network topologies, whether that be on fixed networks with variable connectivity \citep{Collyer2022ACDG}, or to networks containing differing numbers of nodes \citep{ApplebaumQlearnanalysis,mern2021reinforcementlearningindustrialcontrol_arxiv, janisch2023nasimemu, janisch2023symbolicrelationaldeepreinforcement, palmer2023deep}. Indeed, the recent CAGE Challenge 4~\citep{cage_challenge_4_announcement} introduced a level of variability in the number of possible user nodes to encourage solutions addressing this problem.

One suggestion~\citep{palmer2023deep} is to use graph neural-networks (GNNs), specifically Graph Attention Networks~\citep{velickovic2018graph} (GATs) to exploit the structural information available in cyber environments as well as imbue any defensive policy with some degree of invariance to the number of nodes in a network. For penetration testing using an autonomous network attacker,~\citet{janisch2023nasimemu} use an `invariant' architecture for the attacking agent's policy. The agent has a shared encoder for every node, and amalgamates information from nodes as they are discovered by the agent. This is similar to the message passing seen in GNN-based policies.
In~\citet{janisch2023symbolicrelationaldeepreinforcement}, the same authors study the use of a GNN-based policy, together with autoregressive policy decomposition for the \textit{SysAdmin} game \citep{Guestrin_2003}, and test the generalisation ability of the policy on networks of variable size.

To the best of our knowledge, the most relevant prior work is~\citet{mern2021reinforcementlearningindustrialcontrol_arxiv, mern_RL_industrial_control_IEEE}, where the authors implement a reinforcement learning environment for the autonomous defence of an industrial control network, and use an attention-based action-value function. Similarly to the entity-based paradigm, they use separate modules to encode different node-types, and share encoder parameters between all nodes of the same type. A global attention layer is then used to pass information between all node embeddings, before action-value estimates are decoded for all actions on all nodes. The parameters are learnt using a version of the Deep Q-Learning algorithm \citep{mnih2013playingatarideepreinforcement, RAINBOW}. This differs from the explicit policy parameterisation used in this work, trained using PPO. An advantage of a policy-based approach is that it allows for a more natural implementation of composite action spaces.

\subsection{Entities and Objects in Reinforcement Learning}
In this section we provide an overview of attempts to leverage the underlying structure of RL environments, as well as practical approaches for dealing with complex environments with potentially variable observation and action spaces, such as in the case of autonomous cyber defence. In the first instance, particularly in the case of factored MDPs \citep{BOUTILIER_factoredMDPs}, the motivation is to find more efficient solution methods by exploiting the `factorised' structure of an environment. More broadly, all of these approaches contribute to solving the problem of compositional generalisation \citep{zhao2022compositionalgeneralizationobjectorientedworld,compositional_generalization_survey_baihan_lin_2023}. That is, training policies that are able to generalise across environments containing different compositions of familiar objects or entities and dynamics. For further reference, \citet{mohan2024structure} provide a comprehensive survey of structure in reinforcement learning.

\subsubsection{AlphaStar and entity-based Reinforcement learning}\label{sec:alphastar}
AlphaStar~\citep{AlphaStarNature, AlphaStarThesis} is an agent developed to play the game Starcraft II at human professional level. Starcraft II is a real-time strategy game in which two opposing players develop a base of operations and direct armies of units with the goal of defeating their opponent. Starcraft is a hard problem, both in terms of the practicalities of agent design and the complexity of strategies, which captures many of the challenges of real-world deployment of autonomous agents. Of particular relevance is that the number and nature of the entities an agent can observe and act upon varies throughout the game. This problem may be similar to that of nodes joining or leaving a network, or more generally the problem of transferring to an unfamiliar network consisting of familiar entities.
Rather than using purely image-based inputs of the game screen as observations, AlphaStar maintains an entity list consisting of every visible object the agent can interact with. This could be army units, buildings, resources, or terrain. Crucially, this list will vary throughout a game, and between games.

Each entity is associated with a one-dimensional vector that summarises the state and properties of the entity. AlphaStar encodes these entity observations through a three-layer Transformer~\citep{Transformers} to learn the relationships between entities. The entity embeddings generated by the Transformer are merged with other scalar and game-map information to form a joint representation for a single time step's observation, which is used as input to an LSTM~\citep{Hochreiter1997} that learns temporal dependencies.

Since the actions available to the agent depend on the entities that are visible, the policy must be equipped to deal with a variable-size action space, and to map information directly from its entity-based observations to the relevant actions associated with each entity. This is achieved through using skip-connections from the entity embeddings to an entity-selection head. This network computes a key for each entity from the provided entity embeddings, which are then used by a recurrent pointer network~\citep{vinyals2015pointer} to select which entities to act on. This pointer network is invariant to the number of entities available at any given time. Similarly, there is a target entity head which works the same way, but selects a single entity, for actions that might first involve selecting entities and then targeting another. In Starcraft, this could be selecting a group of friendly units to target an opponent.

\subsubsection{Factored MDPs}\label{sec:Factored_MDPs}
Factored MDPs \citep{BOUTILIER_factoredMDPs, Guestrin_2003, Osband2014Factored} provide a framework that allows for more efficient learning on environments that may be naturally `factorised' into distinct components. Here, the reward and transition functions can also be factorised according to these components, and display conditional independence properties. \citet{Guestrin_2003} introduced efficient solution methods for reinforcement learning in factored MDPs as well as a toy environment, \textit{SysAdmin}, on which to study them. In this environment, a system administrator must maintain a network of computers, each of which can be functional or failed. The administrator must decide whether to reboot one of the machines or do nothing at each time step, aiming to maximise rewards received for working machines, while considering that machine failures are influenced by the state of neighbouring machines. In this environment, the transition of each individual machine is conditionally independent of all other machines given its immediate neighbours, hence it can be cast as a factored MDP. The \textit{SysAdmin} example bears resemblance to autonomous cyber defence environments, and the F-MDP framework could be a suitable candidate to express network security problems such as the ones considered in this work.

\subsubsection{Relational Reinforcement Learning}
Relational reinforcement learning (RRL) is the study of solution methods for environments where states and actions can be naturally represented using relational and logical structures \citep{Dzeroski2001_relational}. This framework aims to address the limitations of traditional RL methods when dealing with structured environments and to enable more efficient generalisation across tasks.

In RRL, the state space, action space, and transition function are represented using relational descriptions. This allows for a more compact and expressive representation of complex environments. The key idea is to learn relational policies that can generalise across objects and relations, rather than learning separate policies for each specific instance \citep{Otterlo2005ASO}.

\citet{zambaldi2018deep} study `deep' relational reinforcement learning, and advocate the use of GNN or attention-based function approximators in order to learn abstract representations of relations between entities in an environment. This is opposed to traditional relational RL, where relations are provided a-priori. They find the use of `relational' attention modules allows an agent to generalise zero-shot to unseen environment configurations. This approach essentially discards the traditional formal solution methods based on logical descriptions of the environment. In other words, it is a purely entity-based approach, where the environment is split into distinct entities, and any relations between them are learnt during training.

Similarly, \citet{janisch2023symbolicrelationaldeepreinforcement} study the use of a GNN-based policy on relational planning domains, including the `\textit{SysAdmin}' environment introduced for the study of factored MDPs by \citet{Guestrin_2003}, as mentioned above. They find their approach demonstrates zero-shot generalisation ability in environments with larger numbers of objects than seen during training, including `\textit{SysAdmin}' networks of variable sizes. Motivated by this, \citet{nyberg2024MPNNS_cage_generalisation} recently applied this GNN-based policy architecture to the CAGE Challenge 2 cyber defence environment \citep{cage_challenge_2}. The authors experiment with modifying the default scenario by adding and removing user-level nodes, and testing the zero-shot generalisation performance of the GNN-based policy on these variants.

\subsubsection{Object-oriented Reinforcement Learning}

Similarly to RRL, Object-Oriented Reinforcement Learning (OORL) \citep{DiukWasser2010_OORL} offers a formalism for representing RL environments consisting of distinct objects, where each object belongs to a class with a particular set of features or attributes. That is, objects that fall within a certain class $c_i \in \mathcal{C} = \{c_1, ..., c_m\}$ possess a particular set of class-specific attributes $\{c_i.a_1,...,c_i.a_k\}$. It is similar to the RRL formalism, but less restrictive in terms of defining the interactions between objects. The `object-oriented' or `object-centric' RL terminology has been adopted in research into visual robotics tasks. Often where the objects in the environment are discovered using an object detection architecture, before being fed into a GNN or Transformer-based policy \citep{janner2018reasoning, mosbach2024soldreinforcementlearningslot, locatello2020objectcentriclearningslotattention, haramati2024entitycentricreinforcementlearningobject}, as opposed to being defined as part of the environment in advance. Similarly to deep RRL, the terminology has survived, but when using solutions based on deep learning with transformers or GNNs, the specifics of the formalism is not particularly relevant. What is consistent is the recognition that partitioning an environment into distinct objects with their own features, and training a permutation invariant set-based architecture across these objects results in more efficient training and generalisation capabilities.

\subsubsection{Graph-based Reinforcement Learning for robotic control}

A number of papers have studied the use of GNN or Transformer-based policies on robotics continuous control environments,
which \citet{hong2022structureaware} refers to as `\textit{Inhomogeneous Multitask Reinforcement Learning}'. Here, the agent is tasked with learning policies across multiple tasks that may have different (inhomogeneous) state and action spaces or dynamics. The goal is to leverage shared knowledge and experience across these tasks to improve learning efficiency and performance. This has been successfully applied to robotics environments~\citep{wang2018nervenet, huang2020policy, kurin2021AMORPHEUS, hong2022structureaware}, where the observation and action space are taken to be nodes in a graph of robot joints. An agent may be trained across different robot morphologies given by different graphs, with the goal that the policy is generalisable across all of these settings. Such tasks can be formalised by considering a set of MDPs $\mathcal{M}^K = \{M_1, \dots, M_K\}$ \citep{kurin2021AMORPHEUS, hong2022structureaware}. The set $\mathcal{M}^K$ is inhomogeneous if there exists $i,j \in [K], i \neq j$ such that $dim(\mathcal{S}_i) \neq dim(\mathcal{S}_j)$ or $dim(\mathcal{A}_i) \neq dim(\mathcal{A}_j)$, where $\mathcal{S}_i$ and $\mathcal{A}_i$ refer to the state and action spaces of MDP $M_i$. The aim is then to construct a policy that maximises the average expected discounted return over all environments in $\mathcal{M}^K$, i.e., $\frac{1}{K} \sum_{i=1}^K \mathbb{E}[\mathcal{R}_i]$, where $\mathcal{R}_i$ denotes expected returns in MDP $M_i$. In order to train such a policy, \citet{wang2018nervenet} propose using GNNs to model the relationships between different components of the state and action spaces across tasks.

\section{Entity-based Reinforcement Learning}
To apply entity-based RL to Yawning Titan, we make use of the main constituents of the Entity Neural Network project of~\citet{winter2023entity}. Namely, the Entity Gym interface \citep{entity-gym}, the RogueNet policy architecture \citep{rogue-net}, and the Entity Neural Network Trainer framework for training agents in Entity Gym environments \citep{enn-trainer}. The following section is a summary of components of these works necessary for later discussion and does not represent an original contribution of this paper. For more details on implementation and architecture, refer to the documentation of the repositories of Entity Gym \citep{entity-gym} and the Entity Neural Network Trainer \citep{enn-trainer}, as well as the blog post \citep{winter2023entity} providing a comprehensive introduction to entity-based RL and the motivation behind the Entity Neural Network project.

\subsection{Entity Gym}\label{sec:entity-gym}
In this section we describe the features of the Entity Gym framework~\citep{winter2023entity, entity-gym} and the choices available for specifying an entity-based environment.

\noindent\textbf{Environment}\\
The \textit{Environment} class in Entity Gym defines a standard interface for entity-based RL. Following a similar structure to Gymnasium~\citep{towers2024gymnasium}, this interface includes methods for defining the observation and action spaces, resetting the environment, processing actions, and generating observations. To create a custom entity-based environment, one inherits from the \textit{Environment} class and implements these core abstract methods. This process involves defining all relevant entity types and their features, as well as the action spaces associated with specific entity types. As we do in this paper, if replicating an existing environment such as Yawning Titan which has a Gym or Gymnasium interface, it is important to ensure that the observations and actions in the Entity Gym interface correspond to those in the original version and that functionality is preserved. In other words, it should be the case that the interface and representation of the environment are different, but the information provided to the agent and the fundamental dynamics of the environment are the same.

\textbf{Entities}\\
Any entity-based environment is initialised with lists of possible entities, each with a specified type. Each entity type is simply defined by a list of named features that are associated with that entity type.

\textbf{Observations}\\
Each observation can contain global features, entity-specific features, action masks, and rewards for the current time step. The entity features are stored in a dictionary mapping entity types to lists of feature arrays, where each element in the list is the feature vector for a particular entity. These lists are permitted to be any length at any given time step, allowing for variable numbers of entities in the environment. Action masks, defined for each action type, specify which entities can perform actions and which actions are available at a particular time step. These masks interact directly with the action space, constraining the set of valid actions for each entity. We do not make use of action masks for our environment, although one might imagine them being useful for a real cyber environment, where access to particular nodes may be prevented.

\textbf{Action Spaces}\\
Three kinds of actions spaces are supported:
\begin{itemize}
    \item `\textit{GlobalCategoricalActionSpace}': For global actions that are not associated with specific entities. 
    \item `\textit{CategoricalActionSpace}': This action space allows the agent to direct a set of entities to execute a discrete action, where a different action may be executed for each entity. Each instantiation of this action space is associated with a particular entity type, or set of entity types.
    \item `\textit{SelectEntityActionSpace}': This action space allows a set of entities to choose another entity to act on. Similarly, each instantiation of this action space is associated with particular actor and actee entity types.
\end{itemize}
For the \textit{GlobalCategoricalActionSpace}, one might assign actions that affect a large number of entities simultaneously, or have no direct effect on the entities in the environment but change the state of the agent. For example, moving the top-down camera in a video game such as StarCraft \citep{AlphaStarNature}, which changes the entities that are visible to the agent. For an autonomous network-based agent, this could be any action that has network-wide implications, or perhaps the decision to intervene in the network if malicious activity is detected~\citep{hammar2022learning}.

For the `non-global' \textit{CategoricalActionSpace}, one might assign actions that are associated with an entity-level effect. In the example of a grid-world game, this could be moving entities in canonical directions. In an autonomous cyber defence environment, this could be node-level actions such as rebooting specific nodes, or terminating specific processes.

\textit{SelectEntity} actions could be used for anything that involves directing one entity to interact with another. In AlphaStar, the agent has a \textit{``Target unit''} head in its action pipeline, which allows collections of entities to apply a previously selected action to another entity. In the cyber-defence setting one might imagine implementing an action that allows a centralised agent to select one node to send keying material, a digital certificate or a backup to another node. Additionally, one might imagine treating services or users on a particular node as entities in themselves, and directing the node to select one of its internal entities to act on (for example, terminating a service). One could also define any defensive agent or node issuing defensive commands as an abstract entity in itself, and treat all defensive actions involving operations on specific nodes as \textit{SelectEntityActions}, where the defensive entity selects other nodes to perform actions on. This can be useful for constructing complex action spaces, in combination with the \textit{Categorical} or \textit{GlobalCategorical} action spaces, where an action type could be chosen separately from the node to act on.\\

\subsection{RogueNet}
RogueNet~\citep{rogue-net} is a Transformer-based policy network implementation, designed to work specifically with Entity Gym compatible environments~\citep{entity-gym}, and uses a specially implemented ragged batch data-type \citep{winter2023raggedbuffer} to deal efficiently with variable length observations as input. It is similar to the entity encoding architecture used by \citet{AlphaStarNature} (see \ref{sec:alphastar}).

\noindent\textbf{Input Representation}\\
RogueNet takes entity-based observations as input, represented as a set of features for each entity. The input is structured as a mapping from entity types to ragged batch objects, allowing for efficient training with batching whilst maintaining support for variable numbers of entities across environment instances. These batches are constructed from the lists of entity features that constitute the observations provided to the agent at each time step. Masks are used to prevent information leakage between time steps and environment instances. Global features, such as network connectivity information, could be incorporated and are appended to each entity's feature set, maintaining a consistent input structure.

\textbf{Entity Embedding}\\
An entity embedding head is initialised for every entity type in the environment. The same embedding parameters are used for all entities of a given type. These heads are single-layer MLPs with input and layer normalisation, and ReLU nonlinearity.

\textbf{Transformer Backbone}\\
After entities are embedded, these embeddings are passed through several layers of Transformer blocks, in our case 2.
Each block comprises self-attention and feedforward sub-layers with residual connections and layer normalisation.

\textbf{Action Heads}\\
RogueNet uses multiple action heads based on the action spaces defined in the environment, with the three types defined in Section \ref{sec:entity-gym}. Both \textit{GlobalCategorical} and \textit{Categorical} actions use the same categorical action head implementation. For global categorical actions, a dummy `global' entity is introduced. This entity's representation is updated in each Transformer layer along with other entities, allowing it to aggregate information from all entities. This action head is implemented as a simple affine layer followed by a softmax, taking as input the entity-embedding after being fed through the Transformer backbone of the policy. For \textit{SelectEntity} actions, the action head computes a query $q$ from the embedding of the actor entity, and key $k$ vectors for each selectable actee entity:
\begin{equation*}
q = W_q h_{\text{actor}}, \quad k_i = W_k h_i
\end{equation*}
where $W_q$ and $W_k$ are learnable weight matrices, $h_{\text{actor}}$ is the representation of the acting entity and $h_i$ are the representations of candidate actee entities. A dot product between these embeddings is then used to create a probability distribution over entities. This is similar in function to pointer networks~\citep{vinyals2015pointer}. That is, the probability of selecting entity $i$ is determined by $\frac{q^T k_i}{\sqrt{d_{qk}}}$ where $d_{qk}$ is the dimension of the query and key vectors, which is passed through a softmax over all eligible entities $i$ to produce valid probabilities.

\textbf{Training and Inference}\\
RogueNet can be trained using policy gradient methods, and we use the version of PPO implemented in its companion Entity Neural Network Trainer package \citep{enn-trainer}, which is itself adapted from CleanRL \citep{huang2022cleanrl}. RogueNet's use of ragged tensors and the attention mechanism enables it to process varying numbers of entities without altering its components or structure. This property allows the architecture to be applied to environments with dynamic entity populations or across multiple environments with different numbers of entities, such as network environments with different numbers of nodes. It is straightforward to deploy a trained policy on any Entity Gym environment containing the same types of entities and action spaces as the environment used for training the policy. For our use-case, this is network environments containing variable numbers of nodes.
\section{Yawning Titan}\label{sec:Yawning_Titan}
\begin{figure*}[t]
    \centering    
    \includegraphics[width=0.98\textwidth]{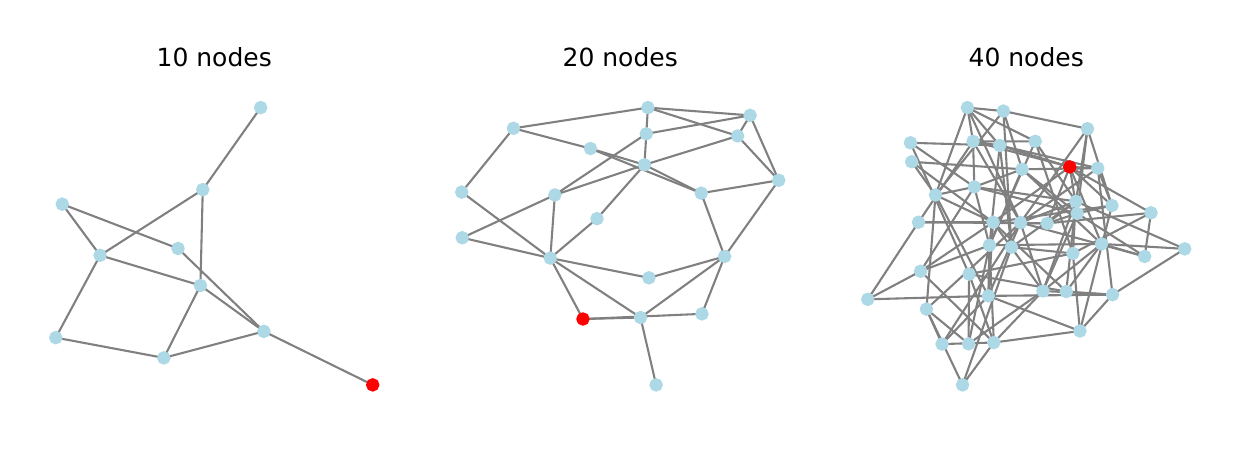}
    \caption{Plots showing examples of the structure of random networks used in the Yawning Titan environment, with entry nodes marked in red.}
    \label{fig:random_network_plots}
\end{figure*}

Yawning Titan is an abstract, graph-based environment designed for the development of RL agents for autonomous cyber defence \citep{Yawning_Titan}. It provides an OpenAI Gym \citep{OpenAIGym} interface for a network-based cyber defence game through which a defending agent (blue agent) can interact with a network environment containing an attacker (red agent). Yawning Titan allows for flexibility in the configuration of the network and the dynamics of the game, including types of nodes, network topology, actions and observations available, termination, rewards, and red agent behaviour. Several works have used and adapted Yawning Titan to train and evaluate autonomous blue agents \citep{Acuto_YT_varied_networks,Yawning_Titan, Collyer2022ACDG} with various different environment configurations. In the following, we describe the main components of the environment configuration we have used for the experiments in this paper, following previous work closely \citep{Collyer2022ACDG} but with some differences to allow for clearer comparisons. Yawning Titan includes more features and options for configuration than discussed here. More information can be found in the documentation and repository, as well as the original paper \citep{Yawning_Titan}.

\subsection{Nodes and Network} \label{sec:netwgen}
The nodes in the environment are simple objects with two main attributes, \textit{Node Vulnerability} and \textit{Compromised Status}. \textit{Node Vulnerability} is a value in the range $[0,1]$ that dictates how easily a node can be compromised by an attacking red agent, with a higher value meaning that the node is more vulnerable. Each node might also have an additional special attribute:
\begin{itemize}
    \item \textbf{High Value / Target Node:} A flag that triggers special environment behaviour when the node is compromised, such as episode termination or a large negative reward for the blue agent.
    \item \textbf{Entry Node:} At the start of an episode, the red agent may only enter the network by attacking designated entry nodes. From there, it can attack nodes adjacent to those it has compromised already.
    \item \textbf{Deceptive Node:} It is possible to configure the environment such that the blue agent may add new deceptive nodes to the network, which divert the red agent away from compromising the true network.
\end{itemize}
For our experiments, we used a single randomly selected \textit{Entry Node}, and we did not use the \textit{High Value Node} and \textit{Deceptive Node} functionality. We mention them here as these are examples of possible varieties of entity type that could be included in more complex environments to make full use of the entity-based framework. Adding the deceptive node functionality would also mean that the number of nodes in the environment could vary during episodes, not just between episodes, as we have tested here. This is something a GNN or Transformer-based policy such as RogueNet should also be robust to in principle.

The network structures underlying each instance of a Yawning Titan environment are constructed using the NetworkX package \citep{hagberg2008exploring}. Following \citet{Collyer2022ACDG}, we randomly generate these networks using the Erd\H{o}s-Renyi model with edge probability equal to $0.1$, with post-processing to make sure the graph is connected. Figure \ref{fig:random_network_plots} shows examples of networks generated in this way for 10, 20 and 40 nodes, which are the network sizes we used for our experiments.

\subsection{Termination}

Both \citet{Yawning_Titan} and \citet{Collyer2022ACDG} consider environment configurations in which episodes terminate either when the red agent compromises a high-value node or when the episode reaches a maximum length. In our case, we fix the episode length at 100 environment steps, with no other termination conditions. This allows us to focus solely on the agent's ability to generalise across different network topologies, without confounding factors from varying episode lengths. 

\subsection{Red Agent}

In cyber defence simulators, the behaviour of the attacking red agent is typically generated in a rules-based fashion, either with some randomisation or entirely deterministic \citep{cage_cyborg_2022, Yawning_Titan, cage_challenge_1, cage_challenge_2}. The red agent is designed to spread through a given network and compromise nodes, while the blue agent is tasked with preventing this from happening. 

For our environment, the red agent has two actions, \textit{Basic Attack} and \textit{Zero Day}, both of which target a single node per time step:
\begin{itemize}
    \item \textbf{Basic Attack} targets a node and, if successful, allows the red agent to compromise it. Success depends on the `skill' level of the red agent, defined as $s \in [0,1]$, and on the vulnerability score $v \in [0,1]$ of the targeted node. Together, skill level and vulnerability define the `attack strength' $a$ as $a = \frac{100s^2}{s + (1-v)}$. The targeted node is compromised if $a > t$, where $t \sim \mathcal{N} (0, 100)$, otherwise it fails.

    \item \textbf{Zero Day} attacks allow the red agent to compromise a targeted node regardless of its own skill level or the vulnerability of the node. This gives the red agent the opportunity to make progress even when facing a strong defensive blue agent. However, so the basic attack is not redundant, this only becomes available once every three environment steps.
    
\end{itemize}

At each environment step, the red agent randomly chooses a node to act on. If the \textit{Zero Day} attack action is available, this will be prioritised and used. Otherwise, the agent must resort to the \textit{Basic Attack}.

Since the focus of our experiments was on varying network topologies, we decided to employ a fixed configuration for the red agent's behaviour. However, the configuration and diversity of the behaviour of the red agent has a large impact on the difficulty of the task the blue agent is faced with, and the policy it will end up learning. For robustness, training a blue agent for real-world deployment would likely involve exposing it to a diverse range of red agent behaviour and configurations, including perhaps an intelligent attacker trained in opposition to the defender \citep{janisch2023nasimemu, hammar2020RLSelfPlayCSLE, msft:cyberbattlesim}.

\subsection{Observations}

The blue agent has access to the compromised status (a Boolean value) and the vulnerability score for every node, with no other information provided. For the agents trained on the standard Gym interface for Yawning Titan, the observation is a vector of these features for all nodes concatenated together. For the Entity Gym interface, the observation is given as a list, where each element of the list is the feature vector for the corresponding node. If there were different node types, the features would be split into separate lists for each node type.

\subsection{Actions}

The blue agent may only choose to act on one node per environment step. Following \citet{Collyer2022ACDG}, we give the blue agent access to two actions: \textit{Reduce Vulnerability} and \textit{Restore Node}. In the standard OpenAI Gym configuration, the action space consists of a combination of each action with each node in the network, amounting to $2n$ actions if $n$ is the number of nodes. 

\begin{itemize}
    \item \textbf{Reduce Vulnerability} allows the blue agent to reduce the vulnerability $v$ of a specific node by a fixed amount. The default amount is $0.2$, with a lower bound on the vulnerability $v$ placed at $0.01$, such that it cannot be reduced further. This is a proactive or preventative action, through which the blue agent can try to anticipate the red agent behaviour and strengthen nodes that are likely to be easily compromised.

    \item \textbf{Restore Node} allows the blue agent to restore a node to its original state. This means that if the node has been compromised, the compromised status will be removed. While, if the vulnerability of the node has changed, it will be reset to its original value. This is a reactive action, through which the blue agent might respond to any inroads the attacker has made into the network.
 \end{itemize}

\subsection{Rewards}
For the reward provided to the blue agent at each time step, we again followed \citet{Collyer2022ACDG} and defined it as the proportion of nodes not compromised at time step $t$: $$R_t = \frac{N - N^c_t}{N}$$ where $R_t$ is the reward at time step $t$, $N$ is the total number of nodes in the environment, and $N^c_t$ is the number of compromised nodes at time step $t$. Defining rewards in this way induces a simple positive reward structure, which encourages the blue agent to minimise the number of compromised nodes. \citet{Collyer2022ACDG} also include a large negative reward of $-100$ for a high-value node being compromised, and a large positive reward of $100$ for an episode reaching its maximum number of steps. We did not make use of these additional rewards in our experiments, since our environment has fixed-length episodes and does not feature high-value nodes. With fixed episodes of length 100 under this reward structure, the maximum reward is 100, with a maximum per-time step reward of 1 if no nodes are compromised.
\section{Entity-based adaptation}
In order to apply entity-based RL to Yawning Titan, we first adapted its underlying simulator to the Entity Gym interface. In this section, we will provide a description of how this can be achieved.
\subsection{Entity Types}
In the Yawning Titan environment, it is possible to make the distinction between different types of nodes. Whilst all nodes behave in fundamentally the same way, there may be some nodes that have a special designation. These are the entry, high-value, and deceptive nodes. For our experiments, we used a single randomly chosen entry node, with all others being generic. For this reason, we have chosen to use a single generic node \textit{Entity Type} in the construction of the entity-based environment. In the RogueNet architecture \citep{rogue-net}, this means that a single shared entity encoder is used for all node embeddings during training.

In more realistic simulators or real networks, it may be beneficial to encode different types of network nodes using different \textit{Entity Type} labels and therefore different encoders. This would allow the policy network to be more expressive and would provide a defensive agent with prior knowledge of the fundamental differences between nodes. It would also be impossible to encode all types of nodes with the same encoder if the features or data available to the agent on each node type differ. For example, in the CybORG environment \citep{standen2021cyborg, cage_challenge_2}, one might imagine designating separate entity types for the \textit{User}, \textit{Enterprise Server} and \textit{Operational Server} nodes.
\subsection{Entity Features}
The blue agent receives observations consisting of a dictionary with keys for each entity type, each with a potentially variable-length list of entity features. For each node, we chose to include the vulnerability score and compromised status. 
\subsection{Action Space}
In Yawning Titan, the blue agent may only take action to intervene on one node per time step. The Entity Gym framework offers three possible types of action, as outlined in Section \ref{sec:entity-gym}. Using the entity-level \textit{CategoricalActionSpace} means that a candidate action is chosen for every entity of a particular type at every environment step. In the case of games containing a large number of simultaneously acting entities like Starcraft, this is necessary to allow the agent to act on any subset of them simultaneously, in a tractable way. In our setting, this would mean that the blue agent could direct any subset of nodes simultaneously to either reset themselves or reduce their vulnerability (with any combination of these actions for the selected nodes). This option may be desirable in real network security applications, but it disrupts the balance of the Yawning Titan game, making it significantly easier for the blue agent. It is likely this would be a similar issue in other autonomous cyber defence game environments. This means that it is necessary to create a structured two-stage action space using a combination of a \textit{SelectEntityActionSpace} and either a \textit{CategoricalActionSpace} or a \textit{GlobalCategoricalActionSpace}.

For selecting nodes to act on, we use a \textit{SelectEntityActionSpace}, which requires defined actor and actee entity types. In order to use this action type, we create a dummy `\textit{Defender}' entity with randomly initialised, learnable features to serve as an actor, which is able to act on all nodes in the network.

For the node-specific actions of \textit{Reduce Vulnerability} and \textit{Restore Node}, these can be implemented using either the `global' categorical action space or the `entity-level' categorical action space. For a \textit{GlobalCategoricalActionSpace}, the RogueNet architecture creates a dummy global entity with randomly initialised learnable features, with information from the nodes of the network shared with its embedding vector through the attention mechanism in the Transformer layers. This single, compressed embedding is then provided as the input for the action head. Combined with the \textit{SelectEntityActionSpace}, this means the agent chooses an `action type' and then a node on which to execute it. Under the entity-level \textit{CategoricalActionSpace}, all embeddings of the nodes are passed into the action head simultaneously in a batch, with a separate action selected for every node independently, based on each node's respective embedding. Conceptually, combining this with a \textit{SelectEntityActionSpace} would be like choosing a node to execute the action that it has selected, rather than choosing the action centrally. In our case, we chose to use the global action space to avoid computing redundant actions. Incidentally, the underlying mechanism is very similar to the approach outlined in \citet{janisch2023symbolicrelationaldeepreinforcement}, where environments are represented as a graph, and an extra `global node' is instantiated to gather global information over a series of GNN message-passing steps. This global node is then used to choose the `action identifier', or action type, before proceeding in an autoregressive fashion.

\citet{WOLK_beyond-cage_generalisation} and \citet{hammar2020RLSelfPlayCSLE} decompose the action space in a similar way, where a distinction is made between selecting a node and selecting an action type. \citet{janisch2023symbolicrelationaldeepreinforcement} generalise this to selecting multiple nodes in the \textit{SysAdmin} game, where each node is selected in sequence dependent on the last. In keeping with the architecture of RogueNet \citep{rogue-net}, we treat each component of the policy independently, similarly to \citet{WOLK_beyond-cage_generalisation}. In other words, both the action type and node are chosen independently by different action heads, with the choices combined to execute the action. This is also similar to the decomposition chosen in \citet{vinyals2017starcraftpolicydecomposition}, and means the policy is simpler to learn in comparison to conditional action sequences. In more complex cyber environments, it might be beneficial to incorporate conditional action selection sequences, like the one used in AlphaStar \citep{AlphaStarNature, AlphaStarThesis} or following the more general framework of conditional action trees proposed in \citet{bamford2021generalisingdiscreteactionspaces}, but we leave this for future research.

\section{Experiments}
In this section, we outline the methodology and present results from the experiments we conducted to compare a baseline MLP-based Proximal Policy Optimization (PPO) agent \citep{schulman2017proximal} with an entity-based RogueNet \citep{winter2023entity} policy, as well as experiments exploring the zero-shot generalisation abilities of entity-based agents in environments with network topologies not seen during training. We drew the baseline PPO agent implementation from Stable Baselines 3 \citep{stable-baselines3}, which is a library that collects stable and performant implementations of popular deep RL algorithms, and is generally reported to have the most consistent performance among open-source PPO implementations~\citep{OpenRLBenchmark,PPOimplementationdetails2022}. We trained the RogueNet agents using its companion Entity Neural Network Trainer package \citep{enn-trainer} which uses a simple implementation of PPO derived from the one provided in the CleanRL library \citep{huang2022cleanrl}. 
\begin{figure*}[t]
    \centering    
    \includegraphics[width=\textwidth]{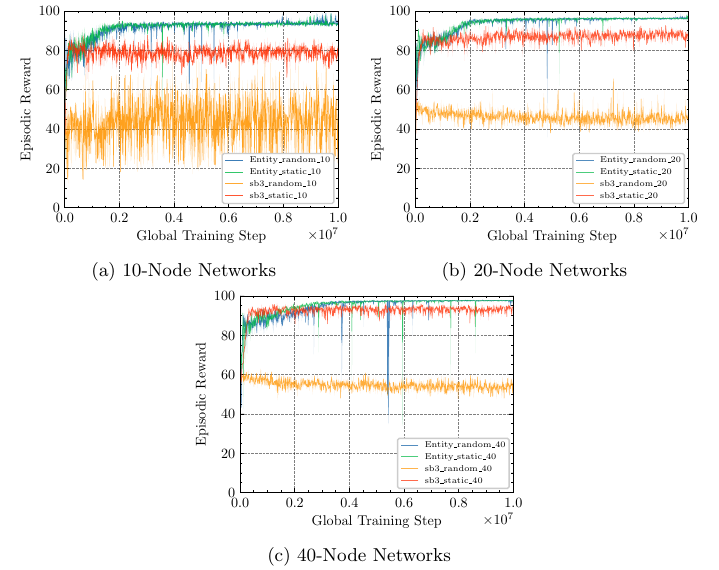}
    \caption{Train-time episodic rewards evaluated on the 10, 20, and 40 node networks respectively. The four agents compared in each evaluation are: baseline PPO agent on the static (\textit{sb3\_static\_[nodes]}) and random (\textit{sb3\_random\_[nodes]}) network environments, and the entity neural network agent on the same static (\textit{sb3\_random\_[nodes]}) and random (\textit{`Entity\_random\_[nodes]'}) environments. Rewards are averaged as the mean over three different random seeds, and shaded error bands are constructed between the maximum and minimum of the three runs. These bands are scarcely visible as there was not a lot of deviation between the three runs.}
    \label{fig:exp_training}
\end{figure*}
\begin{figure*}[t]
    \centering
    \includegraphics[width=\textwidth]{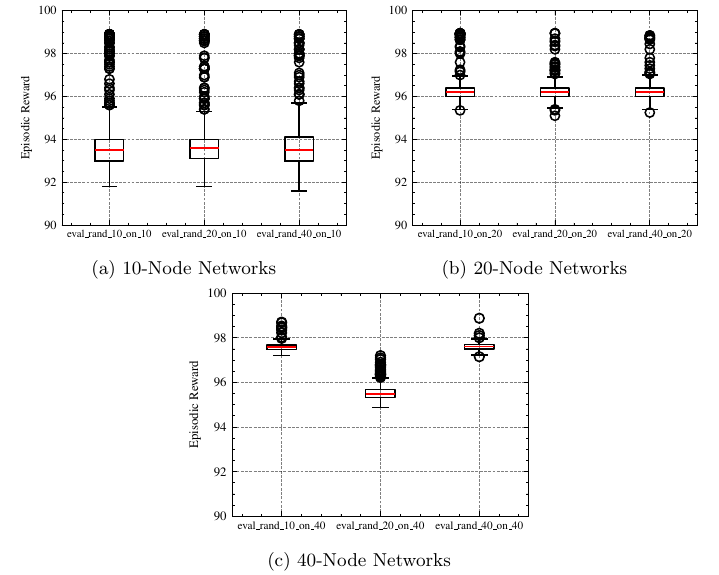}
    \caption{Box-plots of episodic rewards generated from the evaluations of three entity-based agents trained on different network sizes, over 1,000 test episodes. Subfigure (a) shows the evaluations of three entity-based agents trained on 10, 20 and 40-node networks respectively, and evaluated at test-time on random 10-node networks (\textit{eval\_rand\_k\_on\_10} with $\mathbf{k \in \{10, 20, 40\}}$). Subfigure (b) shows the same agents evaluated on 20-node networks (\textit{eval\_rand\_k\_on\_20} with $\mathbf{k \in \{10, 20, 40\}}$). Subfigure (c) shows the same agents evaluated on 40 node networks (\textit{eval\_rand\_k\_on\_40} with $\mathbf{k \in \{10, 20, 40\}}$).}
    \label{fig:exp_eval}
\end{figure*}
\begin{table*}[t]
    \centering
    \begin{adjustbox}{max width=\textwidth}
    \begin{tabular}{|lccccccc|}
        \hline
        & \textbf{mean} & \textbf{std} & \textbf{min} & \textbf{25\%} & \textbf{50\%} & \textbf{75\%} & \textbf{max} \\
        \hline
        \textbf{eval\_rand\_10\_on\_10} & 93.840246 & 1.522998 & 91.000046 & 93.000046 & 93.500053 & 94.000046 & 98.900002 \\
        \textbf{eval\_rand\_20\_on\_10} & 93.764247 & 1.229934 & 91.900063 & 93.100052 & 93.500053 & 94.000046 & 98.900002 \\
        \textbf{eval\_rand\_40\_on\_10} & 92.848245 & 7.654986 & 16.100031 & 93.000053 & 93.500046 & 94.000046 & 98.900002 \\
        \hline
        \textbf{eval\_rand\_10\_on\_20} & 96.278664 & 0.547026 & 95.349945 & 95.999968 & 96.199966 & 96.399963 & 98.949997 \\
        \textbf{eval\_rand\_20\_on\_20} & 96.232763 & 0.414635 & 95.099945 & 95.999960 & 96.199959 & 96.399956 & 98.949997 \\
        \textbf{eval\_rand\_40\_on\_20} & 96.254164 & 0.453860 & 95.249954 & 95.999954 & 96.199974 & 96.399963 & 98.849991 \\
        \hline
        \textbf{eval\_rand\_10\_on\_40} & 97.591049 & 0.187861 & 97.199936 & 97.474945 & 97.574944 & 97.674957 & 98.974998 \\
        \textbf{eval\_rand\_20\_on\_40} & 95.498728 & 0.332559 & 94.874931 & 95.274940 & 95.449917 & 95.624941 & 97.199944 \\
        \textbf{eval\_rand\_40\_on\_40} & 97.601449 & 0.161840 & 97.224937 & 97.499947 & 97.599945 & 97.699951 & 98.875000 \\
        \hline
    \end{tabular}
    \end{adjustbox}
    \caption{Summary statistics for zero-shot episodic evaluation of trained entity-based agents on 1000 random networks of varying sizes. We evaluated 3 policies, trained on network sizes of 10, 20 and 40 respectively, on random networks of 10, 20, and 40 nodes, to assess their generalisation performance. In this table \textit{eval\_rand\_n\_on\_k} refers to the entity-based policy trained on random n-node networks, evaluated on 1000 k-node networks.}
    \label{tab:evaluation_stats}
\end{table*}
\subsection{Training}

All of our experiments employed the Yawning Titan configuration described in Section \ref{sec:Yawning_Titan}, with variable randomised network topologies of 10, 20 and 40 nodes generated via the Erd\H{o}s-Renyi method. We trained each agent on 10,000,000 total environment steps, which amounts to 10,000 episodes with a fixed length of 100 time steps per episode. During training, we evaluated each policy every 10,000 environment steps and logged the resulting episodic reward to produce training graphs.\\\\
\textbf{Exchangeability.} \\
As suggested in \citet{mern_object_exchangeability_paper, mern_object_exchangeability_abstract}, one advantage of the entity-based approach and using a Transformer policy is `object exchangeability'. That is, the robustness of Transformers and GNNs to different permutations of objects in the observation vector or action space without the effectiveness of the policy being impacted. To demonstrate this property, we consider two different training settings, which we refer to as \textit{Static} and \textit{Random}. In the \textit{Static} setting, a single network is randomly initialised at the start of training, and remains the same for every episode; the only thing that varies is the identity of the entry node. In the \textit{Random} setting, an entirely new network is initialised for each episode. The aim of this setup is to evaluate policy robustness across inhomogeneous networks in the training phase, and is similar to the method used by \citet{hong2022structureaware} in `\textit{Multitask Inhomogeneous Reinforcement Learning}' for robotic control.\\\\
\newpage
\textbf{Hyperparameters}\\
To produce the training graphs, we used existing PPO hyperparameter configurations specified in the libraries for both agents. It should be noted that the learning rate used for the optimiser for the entity-based policy was 0.005, compared to 0.0003 in Stable Baselines 3. Different choices of learning rate are likely to affect the gap between asymptotic reward of the two varieties of policy in the \textit{Static} training regime. The full configuration files for the training hyperparameters used can be found in the accompanying repository. A more rigorous analysis would involve hyperparameter sweeps for both varieties of agent, but we believe the fundamental conclusions around the performance difference between \textit{Random} and \textit{Static} modes, as well as the generalisability of entity-based agents, are robust. \\

In total, we trained policies from 3 random seeds for each combination of policy type and environment setting. Considering all the combinations of agents, network sizes and `\textit{Random}' or `\textit{Static}' environments, we have trained 36 policies. Each policy was initialised and trained using a different random seed.
\begin{itemize}
    \item \textbf{sb3\_static\_$n$\_node\_network} $\times 3$: Three different MLP policies trained using the Stable Baselines 3 PPO implementation, on static $n$-node networks randomly generated at the start of each training run, for $n \in (10, 20, 40)$ (the \textit{Static} environment).

    \item \textbf{sb3\_random\_$n$\_node\_network} $\times 3$: Three different MLP policies trained using the Stable Baselines 3 PPO implementation, on $n$-node networks randomly generated with each episode reset for $n \in (10, 20, 40)$ (the \textit{Random} environment).
    
    \item \textbf{EntityYT\_static\_$n$\_node\_network} $\times 3$: Three different Entity Neural Network RogueNet policies, trained using the entity-based version of Yawning Titan, on static $n$-node networks randomly generated at the start of each training run, for $n \in (10, 20, 40)$ (the \textit{Static} environment).
    
    \item \textbf{EntityYT\_random\_$n$\_node\_network} $\times 3$: Three different Entity Neural Network RogueNet policies, trained using the entity-based version of Yawning Titan, on $n$-node networks randomly generated with each episode reset for $n \in (10, 20, 40)$ (the \textit{Random} environment).
\end{itemize}

Figure~\ref{fig:exp_training}(a) shows the reward trajectories of all four types of policies trained on 10-node networks. Both entity-based agents (\textit{`Entity\_random\_10'} and \textit{`Entity\_static\_10'}) show similar reward trajectories that converge rapidly and plateau close to the maximum possible reward of 100. The Stable Baselines 3 PPO agents trained on the single static network (\textit{`sb3\_static\_10'}) converged quickly as well, to a slightly lower asymptotic reward. In line with expectations, the baseline PPO agents trained on a series of random networks that vary across episodes (\textit{`sb3\_random\_10'}) struggled to learn a good policy and displayed high variance in their performance. This supports the notion that a Transformer entity-based policy is more robust to exchangeable representations, and performs no worse than when trained on a static environment, whereas a fixed MLP-based policy struggles to learn an effective policy when the object that each element of the agent's observation vector represents changes with each episode~\citep{mern_object_exchangeability_paper, mern_object_exchangeability_abstract, zhao2022compositionalgeneralizationobjectorientedworld}.

Figures~\ref{fig:exp_training}(b) and (c) show similar results on 20-node and 40-node networks, respectively. Here, the difference between the asymptotic episodic reward of the policies trained in the \textit{Static} regime grows smaller with a larger number of nodes. This may be, in part,  due to the nature of the reward function defined as the proportion of non-compromised nodes at time $t$. Since the skill and behaviour of the red agent are unchanged, this naturally means that a lower proportion of nodes will be compromised on a given time step, and thus rewards for the blue agent will be higher in larger environments. Nevertheless, the Stable Baselines 3 agent trained on randomly generated networks still does comparatively poorly, and struggles to learn a policy that is invariant to different network topologies.

\subsection{Generalisation to unseen topologies}

The second set of experiments we conducted was aimed at evaluating how well entity-based agents generalise at test-time to network sizes not seen during training. As the baseline MLP policies do not support dynamically sized inputs that are different from the ones used at train-time, they cannot be evaluated on varying network topologies at test-time.

To assess the generalisation ability of entity-based policies across different network sizes, we took three policies trained in the \textit{Random} training regime, one for each network size of 10, 20 and 40. We then evaluated each of the policies on random networks for each of the network sizes. This means that we evaluated a policy trained on random 10-node networks in the same context of random 10-node networks but also on 20 and 40-node networks (`out-of-training' context). All evaluations were carried out over $1000$ episodes and so $1000$ different random networks.

Figure \ref{fig:exp_eval} displays the distribution of episodic rewards over 1,000 test episodes in the form of box plots, obtained from the evaluations of the three RogueNet agents trained on different network sizes. Subfigure \ref{fig:exp_eval}(a) presents evaluations of agents trained on 10, 20 and 40-node networks respectively, and evaluated on random 10-node networks at test-time (\textit{eval\_rand\_k\_on\_10} with $\mathbf{k \in \{10, 20, 40\}}$). Subfigures \ref{fig:exp_eval}(b) and \ref{fig:exp_eval}(c) present the corresponding evaluations on 20-node networks, \ref{fig:exp_eval}(b), and 40-node networks, \ref{fig:exp_eval}(c). Summary statistics of the same experiments are presented in Table \ref{tab:evaluation_stats}.

Overall, we can observe a negligible difference in the performance of the three entity-based agents across the different network sizes when evaluated at test-time.  For example, the agent trained on 40-node networks performs just as well when evaluated on 10-node networks as the agent that was trained natively on 10-node networks. In fact, it shows slightly lower variance in episodic reward than the agent trained on 10 nodes, with $0.87$ standard deviation against the $1.31$ standard deviation of the native 10-node agent. The only exception is the agent trained on 20-node networks and evaluated on 40-node networks, which shows slightly lower mean reward compared to the other two agents. As can be seen from the summary statistics results in Table \ref{tab:evaluation_stats}, this case presents a higher standard deviation in the rewards distribution (0.332) compared to the other two cases (0.187 and 0.161). These results suggest the ability of an entity-based agent to learn a general policy that performs consistently on any network size. The generality of these results would need to be verified in more complex environment settings or simulators, and different varieties of network topologies.

\section{Discussion}
In this paper, we have demonstrated the basic functionality and the strong generalisation capabilities of an entity-based RL method through the use of the Entity Gym framework for the Yawning Titan autonomous cyber defence environment. We suggest that other, similar or more complex cyber defence simulators could be readily adapted in the same way, due to the fundamental node-based structure of the observation and action spaces. In fact, when designing a simulator, it is likely to be much more straightforward to use an interface such as Entity Gym than devising rules for how a fixed-input agent can interact through the standard Gymnasium interface. One drawback is the lack of out-of-the-box implementations of algorithms designed to work with entity-based environments, and tuned for optimal performance with complex Transformer or GNN policies. Indeed, as can be seen from the training plots, the RogueNet policy displayed occasional instability during training in the environment setting and hyperparameters we used, although this might be rectifiable by using a lower learning rate. We believe any shortcomings in out-of-the-box performance of an entity-based policy is outweighed by the potential for training and generalisation across variable network topologies, and can be rectified with more development and testing. We suggest that future cyber defence environments be designed using an entity-based philosophy, and constructed with support for using an interface akin to Entity Gym.

\section{Future work}
\subsection{Network and Global Information}
If the network structure is known to the defender, it could be beneficial to provide this information to the blue agent. This could potentially be in the form of incorporating structural bias into the policy network by using a GNN or Graph Attention Network \citep{palmer2023deep}. In the RogueNet architecture, introducing attention masking over node connections to reflect the structure of the network would be a relatively straightforward change.

Alternatively, as in continuous control~\citep{kurin2021AMORPHEUS}, if it is detrimental to restrict message-passing to neighbouring nodes, it is also possible to include global features that contain network connectivity information. Similarly to~\citet{hong2022structureaware}, one might use a full Transformer policy, but with a graph-based positional encoding scheme.

\subsection{Environment complexity and intra-episode variation}
We have demonstrated the ability of the entity-based framework and a Transformer-based policy to train on varying network topologies with constant nodes, as well as to generalise zero-shot to network sizes not seen during training. It is also straightforward to train a policy on varying numbers of nodes during training, where the agent sees a different-sized network with every episode. The agent performs well and also converges to a high reward, although we do not include this here since the reward function is dependent in-part on the size of the network. In the future, one might expect this kind of variety to be important for training robust agents.

Another important benefit of the Entity Gym, and entity-based or object-oriented RL more generally, is the support for natural variation in the population of entities within an episode, not just between episodes. In a deployment on a real network, the number of nodes observed by a defensive agent may vary over time. This could be due to devices dropping in and out of a wireless connection, being manually connected or disconnected by administrators, or communication being blocked by an adversary. In the Yawning Titan environment, it is possible to enable the use of deceptive nodes by the blue agent, which would allow the agent to add new nodes to the environment mid-episode. 

\section{Acknowledgments}
The authors would like to acknowledge helpful feedback and comments from Liz Bates and Ed Chapman. This work was supported by the Defence and Security Programme at The Alan Turing Institute, funded by the Government Communications Headquarters (GCHQ). Alberto Caron was funded by the Defence Science and Technology Laboratory (DSTL) which is an executive agency of the UK Ministry of Defence providing world-class expertise and delivering cutting-edge science and technology for the benefit of the nation and allies.

\newpage
\bibliographystyle{apalike}
\bibliography{references}

\begin{thebibliography}{}

\bibitem[Acuto and Maskell, 2023]{Acuto_YT_varied_networks}
Acuto, A. and Maskell, S. (2023).
\newblock Defending the unknown: Exploring reinforcement learning agents' deployment in realistic, unseen networks.

\bibitem[Andrew et~al., 2022]{Yawning_Titan}
Andrew, A., Spillard, S., Collyer, J., and Dhir, N. (2022).
\newblock Developing optimal causal cyber-defence agents via cyber security simulation.

\bibitem[Applebaum et~al., 2022]{ApplebaumQlearnanalysis}
Applebaum, A., Dennler, C., Dwyer, P., Moskowitz, M., Nguyen, H., Nichols, N., Park, N., Rachwalski, P., Rau, F., Webster, A., and Wolk, M. (2022).
\newblock Bridging automated to autonomous cyber defense: Foundational analysis of tabular q-learning.
\newblock In {\em Proceedings of the 15th ACM Workshop on Artificial Intelligence and Security}, AISec'22, page 149–159, New York, NY, USA. Association for Computing Machinery.

\bibitem[Bamford and Ovalle, 2021]{bamford2021generalisingdiscreteactionspaces}
Bamford, C. and Ovalle, A. (2021).
\newblock Generalising discrete action spaces with conditional action trees.
\newblock {\em 2021 IEEE Conference on Games (CoG)}, pages 1--8.

\bibitem[Bates et~al., 2023]{bates2023reward}
Bates, E., Mavroudis, V., and Hicks, C. (2023).
\newblock Reward shaping for happier autonomous cyber security agents.
\newblock In {\em Proceedings of the 16th ACM Workshop on Artificial Intelligence and Security}, pages 221--232.

\bibitem[Bello et~al., 2017]{bello2017neuralcombinatorialoptimizationreinforcement}
Bello, I., Pham, H., Le, Q.~V., Norouzi, M., and Bengio, S. (2017).
\newblock Neural combinatorial optimization with reinforcement learning.
\newblock In {\em 5th International Conference on Learning Representations, {ICLR} 2017, Toulon, France, April 24-26, 2017, Workshop Track Proceedings}. OpenReview.net.

\bibitem[Boutilier et~al., 2000]{BOUTILIER_factoredMDPs}
Boutilier, C., Dearden, R., and Goldszmidt, M. (2000).
\newblock Stochastic dynamic programming with factored representations.
\newblock {\em Artificial Intelligence}, 121(1):49--107.

\bibitem[Brockman et~al., 2016]{OpenAIGym}
Brockman, G., Cheung, V., Pettersson, L., Schneider, J., Schulman, J., Tang, J., and Zaremba, W. (2016).
\newblock Openai gym.

\bibitem[Choi, 2020]{AlphaStarThesis}
Choi, D. (2020).
\newblock Alphastar: Considerations and human-like constraints for deep learning game interfaces.
\newblock Master's thesis.

\bibitem[Collyer et~al., 2022]{Collyer2022ACDG}
Collyer, J., Andrew, A., and Hodges, D. (2022).
\newblock {ACD-G: Enhancing Autonomous Cyber Defense Agent Generalization through Graph Embedded Network Representation}.
\newblock In {\em Proceedings of the 39th International Conference on Machine Learning (ICML 2022), ML4Cyber Workshop}, Baltimore, Maryland, USA.
\newblock 17-23 July 2022.

\bibitem[Diuk et~al., 2008]{Diuk_Object_Oriented_MDP_paper}
Diuk, C., Cohen, A., and Littman, M.~L. (2008).
\newblock An object-oriented representation for efficient reinforcement learning.
\newblock In {\em Proceedings of the 25th International Conference on Machine Learning}, ICML '08, page 240–247, New York, NY, USA. Association for Computing Machinery.

\bibitem[Džeroski et~al., 2001]{Dzeroski2001_relational}
Džeroski, S., De~Raedt, L., and Driessens, K. (2001).
\newblock Relational reinforcement learning.
\newblock {\em Machine Learning}, 43(1):7--52.

\bibitem[Foley et~al., 2022a]{foley2022autonomous}
Foley, M., Hicks, C., Highnam, K., and Mavroudis, V. (2022a).
\newblock Autonomous network defence using reinforcement learning.
\newblock In {\em Proceedings of the 2022 ACM on Asia Conference on Computer and Communications Security}, pages 1252--1254.

\bibitem[Foley et~al., 2022b]{foley2023inroadsautonomousnetworkdefence}
Foley, M., Wang, M., M, Z., Hicks, C., and Mavroudis, V. (2022b).
\newblock Inroads into autonomous network defence using explained reinforcement learning.
\newblock In {\em Proceedings of the Conference on Applied Machine Learning in Information Security, {CAMLIS} 2022, Arlington, Virginia, USA, October 20-21, 2022}, volume 3391 of {\em {CEUR} Workshop Proceedings}, pages 1--19. CEUR-WS.org.

\bibitem[Guestrin et~al., 2003]{Guestrin_2003}
Guestrin, C., Koller, D., Parr, R., and Venkataraman, S. (2003).
\newblock Efficient solution algorithms for factored mdps.
\newblock {\em Journal of Artificial Intelligence Research}, 19:399–468.

\bibitem[Hagberg et~al., 2008]{hagberg2008exploring}
Hagberg, A.~A., Schult, D.~A., and Swart, P.~J. (2008).
\newblock Exploring network structure, dynamics, and function using networkx.
\newblock In {\em Proceedings of the 7th Python in Science Conference (SciPy2008)}, pages 11--15, Pasadena, CA USA.

\bibitem[Hammar and Stadler, 2020]{hammar2020RLSelfPlayCSLE}
Hammar, K. and Stadler, R. (2020).
\newblock Finding effective security strategies through reinforcement learning and self-play.
\newblock In {\em 2020 16th International Conference on Network and Service Management (CNSM)}, pages 1--9.

\bibitem[Hammar and Stadler, 2022]{hammar2022learning}
Hammar, K. and Stadler, R. (2022).
\newblock Learning security strategies through game play and optimal stopping.
\newblock 17-23 July 2022.

\bibitem[Haramati et~al., 2024]{haramati2024entitycentricreinforcementlearningobject}
Haramati, D., Daniel, T., and Tamar, A. (2024).
\newblock Entity-centric reinforcement learning for object manipulation from pixels.
\newblock In {\em The Twelfth International Conference on Learning Representations}.

\bibitem[Hessel et~al., 2017]{RAINBOW}
Hessel, M., Modayil, J., van Hasselt, H., Schaul, T., Ostrovski, G., Dabney, W., Horgan, D., Piot, B., Azar, M., and Silver, D. (2017).
\newblock Rainbow: Combining improvements in deep reinforcement learning.

\bibitem[Hicks et~al., 2023]{canaries2023}
Hicks, C., Mavroudis, V., Foley, M., Davies, T., Highnam, K., and Watson, T. (2023).
\newblock Canaries and whistles: Resilient drone communication networks with (or without) deep reinforcement learning.
\newblock In {\em Proceedings of the 16th ACM Workshop on Artificial Intelligence and Security}, AISec '23, page 91–101, New York, NY, USA. Association for Computing Machinery.

\bibitem[Hochreiter and Schmidhuber, 1997]{Hochreiter1997}
Hochreiter, S. and Schmidhuber, J. (1997).
\newblock Long short-term memory.
\newblock {\em Neural Computation}, 9(8):1735--1780.

\bibitem[Hong et~al., 2022]{hong2022structureaware}
Hong, S., Yoon, D., and Kim, K.-E. (2022).
\newblock Structure-aware transformer policy for inhomogeneous multi-task reinforcement learning.
\newblock In {\em International Conference on Learning Representations}.

\bibitem[Huang et~al., 2022a]{PPOimplementationdetails2022}
Huang, S., Dossa, R. F.~J., Raffin, A., Kanervisto, A., and Wang, W. (2022a).
\newblock The 37 implementation details of proximal policy optimization.
\newblock In {\em ICLR Blog Track}.
\newblock https://iclr-blog-track.github.io/2022/03/25/ppo-implementation-details/.

\bibitem[Huang et~al., 2022b]{huang2022cleanrl}
Huang, S., Dossa, R. F.~J., Ye, C., Braga, J., Chakraborty, D., Mehta, K., and Araújo, J.~G. (2022b).
\newblock Cleanrl: High-quality single-file implementations of deep reinforcement learning algorithms.
\newblock {\em Journal of Machine Learning Research}, 23(274):1--18.

\bibitem[Huang et~al., 2024]{OpenRLBenchmark}
Huang, S., Gallouédec, Q., Felten, F., Raffin, A., Dossa, R. F.~J., Zhao, Y., Sullivan, R., Makoviychuk, V., Makoviichuk, D., Danesh, M.~H., Roumégous, C., Weng, J., Chen, C., Rahman, M.~M., Araújo, J. G.~M., Quan, G., Tan, D., Klein, T., Charakorn, R., Towers, M., Berthelot, Y., Mehta, K., Chakraborty, D., KG, A., Charraut, V., Ye, C., Liu, Z., Alegre, L.~N., Nikulin, A., Hu, X., Liu, T., Choi, J., and Yi, B. (2024).
\newblock Open rl benchmark: Comprehensive tracked experiments for reinforcement learning.

\bibitem[Huang et~al., 2020]{huang2020policy}
Huang, W., Mordatch, I., and Pathak, D. (2020).
\newblock One policy to control them all: shared modular policies for agent-agnostic control.
\newblock In {\em Proceedings of the 37th International Conference on Machine Learning}, ICML'20. JMLR.org.

\bibitem[Janisch et~al., 2023a]{janisch2023nasimemu}
Janisch, J., Pevn{\'{y}}, T., and Lis{\'{y}}, V. (2023a).
\newblock Nasimemu: Network attack simulator {\&} emulator for training agents generalizing to novel scenarios.
\newblock In {\em Computer Security. {ESORICS} 2023 International Workshops - CPS4CIP, ADIoT, SecAssure, WASP, TAURIN, PriST-AI, and SECAI, The Hague, The Netherlands, September 25-29, 2023, Revised Selected Papers, Part {II}}, volume 14399 of {\em Lecture Notes in Computer Science}, pages 589--608. Springer.

\bibitem[Janisch et~al., 2023b]{janisch2023symbolicrelationaldeepreinforcement}
Janisch, J., Pevný, T., and Lisý, V. (2023b).
\newblock Symbolic relational deep reinforcement learning based on graph neural networks and autoregressive policy decomposition.

\bibitem[Janner et~al., 2019]{janner2018reasoning}
Janner, M., Levine, S., Freeman, W.~T., Tenenbaum, J.~B., Finn, C., and Wu, J. (2019).
\newblock Reasoning about physical interactions with object-centric models.
\newblock In {\em International Conference on Learning Representations}.

\bibitem[Kurin et~al., 2021]{kurin2021AMORPHEUS}
Kurin, V., Igl, M., Rockt{\"a}schel, T., Boehmer, W., and Whiteson, S. (2021).
\newblock My body is a cage: the role of morphology in graph-based incompatible control.
\newblock In {\em International Conference on Learning Representations}.

\bibitem[Lin et~al., 2023]{compositional_generalization_survey_baihan_lin_2023}
Lin, B., Bouneffouf, D., and Rish, I. (2023).
\newblock A survey on compositional generalization in applications.

\bibitem[Locatello et~al., 2020]{locatello2020objectcentriclearningslotattention}
Locatello, F., Weissenborn, D., Unterthiner, T., Mahendran, A., Heigold, G., Uszkoreit, J., Dosovitskiy, A., and Kipf, T. (2020).
\newblock Object-centric learning with slot attention.

\bibitem[Mambelli et~al., 2022]{mambelli2022compositionalmultiobjectreinforcementlearning}
Mambelli, D., Träuble, F., Bauer, S., Schölkopf, B., and Locatello, F. (2022).
\newblock Compositional multi-object reinforcement learning with linear relation networks.

\bibitem[Mern et~al., 2021]{mern2021reinforcementlearningindustrialcontrol_arxiv}
Mern, J., Hatch, K., Silva, R., Brush, J., and Kochenderfer, M.~J. (2021).
\newblock Reinforcement learning for industrial control network cyber security orchestration.

\bibitem[Mern et~al., 2022]{mern_RL_industrial_control_IEEE}
Mern, J., Hatch, K., Silva, R., Hickert, C., Sookoor, T., and Kochenderfer, M.~J. (2022).
\newblock Autonomous attack mitigation for industrial control systems.
\newblock In {\em 2022 52nd Annual IEEE/IFIP International Conference on Dependable Systems and Networks Workshops (DSN-W)}, pages 28--36.

\bibitem[Mern et~al., 2019]{mern_object_exchangeability_abstract}
Mern, J., Sadigh, D., and Kochenderfer, M. (2019).
\newblock Object exchangeability in reinforcement learning: Extended abstract.

\bibitem[Mern et~al., 2020]{mern_object_exchangeability_paper}
Mern, J., Sadigh, D., and Kochenderfer, M.~J. (2020).
\newblock Exchangeable input representations for reinforcement learning.

\bibitem[{Microsoft Defender Research Team}, 2021]{msft:cyberbattlesim}
{Microsoft Defender Research Team} (2021).
\newblock Cyberbattlesim.
\newblock \url{https://github.com/microsoft/cyberbattlesim}.
\newblock Created by Christian Seifert, Michael Betser, William Blum, James Bono, Kate Farris, Emily Goren, Justin Grana, Kristian Holsheimer, Brandon Marken, Joshua Neil, Nicole Nichols, Jugal Parikh, Haoran Wei.

\bibitem[Mnih et~al., 2013]{mnih2013playingatarideepreinforcement}
Mnih, V., Kavukcuoglu, K., Silver, D., Graves, A., Antonoglou, I., Wierstra, D., and Riedmiller, M. (2013).
\newblock Playing atari with deep reinforcement learning.

\bibitem[Mohan et~al., 2024]{mohan2024structure}
Mohan, A., Zhang, A., and Lindauer, M. (2024).
\newblock Structure in deep reinforcement learning: A survey and open problems.
\newblock {\em Journal of Artificial Intelligence Research}, 79:1167--1236.

\bibitem[Molina-Markham et~al., 2021]{FARLAND}
Molina-Markham, A., Miniter, C., Powell, B., and Ridley, A. (2021).
\newblock Network environment design for autonomous cyberdefense.

\bibitem[Mosbach et~al., 2024]{mosbach2024soldreinforcementlearningslot}
Mosbach, M., Ewertz, J.~N., Villar-Corrales, A., and Behnke, S. (2024).
\newblock Sold: Reinforcement learning with slot object-centric latent dynamics.

\bibitem[Nguyen and Reddi, 2021]{nguyen2021deep}
Nguyen, T.~T. and Reddi, V.~J. (2021).
\newblock Deep reinforcement learning for cyber security.
\newblock {\em IEEE Transactions on Neural Networks and Learning Systems}, 34(8):3779--3795.

\bibitem[Nyberg and Johnson, 2024]{nyberg2024MPNNS_cage_generalisation}
Nyberg, J. and Johnson, P. (2024).
\newblock Structural generalization in autonomous cyber incident response with message-passing neural networks and reinforcement learning.

\bibitem[Osband and Van~Roy, 2014]{Osband2014Factored}
Osband, I. and Van~Roy, B. (2014).
\newblock Near-optimal reinforcement learning in factored mdps.
\newblock In Ghahramani, Z., Welling, M., Cortes, C., Lawrence, N., and Weinberger, K., editors, {\em Advances in Neural Information Processing Systems}, volume~27. Curran Associates, Inc.

\bibitem[Palmer et~al., 2023]{palmer2023deep}
Palmer, G., Parry, C., Harrold, D. J.~B., and Willis, C. (2023).
\newblock Deep reinforcement learning for autonomous cyber operations: A survey.

\bibitem[Raffin et~al., 2021]{stable-baselines3}
Raffin, A., Hill, A., Gleave, A., Kanervisto, A., Ernestus, M., and Dormann, N. (2021).
\newblock Stable-baselines3: Reliable reinforcement learning implementations.
\newblock {\em Journal of Machine Learning Research}, 22(268):1--8.

\bibitem[Roy et~al., 2021]{Roy_2021}
Roy, R., Raiman, J., Kant, N., Elkin, I., Kirby, R., Siu, M., Oberman, S., Godil, S., and Catanzaro, B. (2021).
\newblock Prefixrl: Optimization of parallel prefix circuits using deep reinforcement learning.
\newblock In {\em 2021 58th ACM/IEEE Design Automation Conference (DAC)}. IEEE.

\bibitem[Schulman et~al., 2017]{schulman2017proximal}
Schulman, J., Wolski, F., Dhariwal, P., Radford, A., and Klimov, O. (2017).
\newblock Proximal policy optimization algorithms.

\bibitem[Standen et~al., 2021a]{cage_challenge_1}
Standen, M., Bowman, D., Hoang, S., Richer, T., Lucas, M., and Tassel, R.~V. (2021a).
\newblock Cyber autonomy gym for experimentation challenge 1.
\newblock \url{https://github.com/cage-challenge/cage-challenge-1}.

\bibitem[Standen et~al., 2022a]{cage_challenge_2}
Standen, M., Bowman, D., Hoang, S., Richer, T., Lucas, M., Tassel, R.~V., Vu, P., and Kiely, M. (2022a).
\newblock Cyber autonomy gym for experimentation challenge 2.
\newblock \url{https://github.com/cage-challenge/cage-challenge-2}.

\bibitem[Standen et~al., 2022b]{cage_cyborg_2022}
Standen, M., Bowman, D., Hoang, S., Richer, T., Lucas, M., Tassel, R.~V., Vu, P., Kiely, M., C., K., Konschnik, N., and Collyer, J. (2022b).
\newblock Cyber operations research gym.
\newblock \url{https://github.com/cage-challenge/CybORG}.

\bibitem[Standen et~al., 2021b]{standen2021cyborg}
Standen, M., Lucas, M., Bowman, D., Richer, T.~J., Kim, J., and Marriott, D. (2021b).
\newblock Cyborg: A gym for the development of autonomous cyber agents.

\bibitem[Towers et~al., 2024]{towers2024gymnasium}
Towers, M., Kwiatkowski, A., Terry, J., Balis, J.~U., De~Cola, G., Deleu, T., Goul{\~a}o, M., Kallinteris, A., Krimmel, M., KG, A., et~al. (2024).
\newblock Gymnasium: A standard interface for reinforcement learning environments.
\newblock {\em arXiv preprint arXiv:2407.17032}.

\bibitem[{TTCP CAGE Working Group}, 2023]{cage_challenge_4_announcement}
{TTCP CAGE Working Group} (2023).
\newblock Ttcp cage challenge 4.
\newblock \url{https://github.com/cage-challenge/cage-challenge-4}.

\bibitem[van Otterlo, 2005]{Otterlo2005ASO}
van Otterlo, M. (2005).
\newblock A survey of reinforcement learning in relational domains.
\newblock {\em CTIT technical report series}.

\bibitem[Vaswani et~al., 2017]{Transformers}
Vaswani, A., Shazeer, N., Parmar, N., Uszkoreit, J., Jones, L., Gomez, A.~N., Kaiser, L.~u., and Polosukhin, I. (2017).
\newblock Attention is all you need.
\newblock In {\em Advances in Neural Information Processing Systems}, volume~30. Curran Associates, Inc.

\bibitem[Velickovic et~al., 2018]{velickovic2018graph}
Velickovic, P., Cucurull, G., Casanova, A., Romero, A., Li{\`{o}}, P., and Bengio, Y. (2018).
\newblock Graph attention networks.
\newblock In {\em 6th International Conference on Learning Representations, {ICLR} 2018, Vancouver, BC, Canada, April 30 - May 3, 2018, Conference Track Proceedings}. OpenReview.net.

\bibitem[Vinyals et~al., 2019]{AlphaStarNature}
Vinyals, O., Babuschkin, I., Czarnecki, W.~M., Mathieu, M., Dudzik, A., Chung, J., Choi, D.~H., Powell, R., Ewalds, T., Georgiev, P., Oh, J., Horgan, D., Kroiss, M., Danihelka, I., Huang, A., Sifre, L., Cai, T., Agapiou, J.~P., Jaderberg, M., Vezhnevets, A.~S., Leblond, R., Pohlen, T., Dalibard, V., Budden, D., Sulsky, Y., Molloy, J., Paine, T.~L., Gulcehre, C., Wang, Z., Pfaff, T., Wu, Y., Ring, R., Yogatama, D., W{\"u}nsch, D., McKinney, K., Smith, O., Schaul, T., Lillicrap, T., Kavukcuoglu, K., Hassabis, D., Apps, C., and Silver, D. (2019).
\newblock Grandmaster level in starcraft ii using multi-agent reinforcement learning.
\newblock {\em Nature}, 575(7782):350--354.

\bibitem[Vinyals et~al., 2017]{vinyals2017starcraftpolicydecomposition}
Vinyals, O., Ewalds, T., Bartunov, S., Georgiev, P., Vezhnevets, A.~S., Yeo, M., Makhzani, A., Küttler, H., Agapiou, J., Schrittwieser, J., Quan, J., Gaffney, S., Petersen, S., Simonyan, K., Schaul, T., van Hasselt, H., Silver, D., Lillicrap, T., Calderone, K., Keet, P., Brunasso, A., Lawrence, D., Ekermo, A., Repp, J., and Tsing, R. (2017).
\newblock Starcraft ii: A new challenge for reinforcement learning.

\bibitem[Vinyals et~al., 2015]{vinyals2015pointer}
Vinyals, O., Fortunato, M., and Jaitly, N. (2015).
\newblock Pointer networks.
\newblock In {\em Advances in Neural Information Processing Systems}, volume~28. Curran Associates, Inc.

\bibitem[Wang et~al., 2018]{wang2018nervenet}
Wang, T., Liao, R., Ba, J., and Fidler, S. (2018).
\newblock Nervenet: Learning structured policy with graph neural networks.
\newblock In {\em International Conference on Learning Representations}.

\bibitem[Wasser, 2010]{DiukWasser2010_OORL}
Wasser, C. G.~D. (2010).
\newblock {\em An Object-Oriented Representation for Efficient Reinforcement Learning}.
\newblock PhD thesis, Rutgers, The State University of New Jersey, New Brunswick, New Jersey.
\newblock A dissertation submitted to the Graduate School-New Brunswick, Rutgers, The State University of New Jersey in partial fulfillment of the requirements for the degree of Doctor of Philosophy, Graduate Program in Computer Science. Written under the direction of Michael L. Littman.

\bibitem[Winter, 2023]{winter2023raggedbuffer}
Winter, C. (2023).
\newblock {ragged-buffer} version 0.3.7.
\newblock \url{https://crates.io/crates/ragged-buffer/0.3.7}.
\newblock Accessed: 2024-03-22.

\bibitem[Winter et~al., 2021]{rogue-net}
Winter, C., Bamford, C., Huang, C., Matricon, T., and Kanervisto, A. (2021).
\newblock Roguenet: Entity gym compatible ragged batch transformer implementation.
\newblock \url{https://github.com/entity-neural-network/rogue-net}.

\bibitem[Winter et~al., 2023a]{winter2023entity}
Winter, C., Bamford, C., Huang, C., Matricon, T., and Kanervisto, A. (2023a).
\newblock Entity-based reinforcement learning.
\newblock {\em Clemens' Blog}.

\bibitem[Winter et~al., 2023b]{entity-gym}
Winter, C., Bamford, C., Huang, C., Matricon, T., and Kanervisto, A. (2023b).
\newblock Entity gym: Standard interface for entity based reinforcement learning environments.
\newblock \url{https://github.com/entity-neural-network/entity-gym}.
\newblock Accessed: 2024-02-28.

\bibitem[Winter et~al., 2023c]{enn-trainer}
Winter, C., Bamford, C., Huang, C., Matricon, T., and Kanervisto, A. (2023c).
\newblock Entity neural network trainer: Reinforcement learning training framework for entity-gym environments.
\newblock \url{https://github.com/entity-neural-network/enn-trainer}.
\newblock Accessed: 2024-07-09.

\bibitem[Wolk et~al., 2022]{WOLK_beyond-cage_generalisation}
Wolk, M., Applebaum, A., Dennler, C., Dwyer, P., Moskowitz, M., Nguyen, H., Nichols, N., Park, N., Rachwalski, P., Rau, F., , and Webster, A. (2022).
\newblock Beyond cage: Investigating generalization of learned autonomous network defense policies.
\newblock In {\em Reinforcement Learning for Real Life (RL4RealLife) Workshop at NeurIPS 2022}.

\bibitem[Zaheer et~al., 2017]{zaheer2017deepsets}
Zaheer, M., Kottur, S., Ravanbakhsh, S., P{\'{o}}czos, B., Salakhutdinov, R., and Smola, A.~J. (2017).
\newblock Deep sets.
\newblock In {\em Advances in Neural Information Processing Systems 30: Annual Conference on Neural Information Processing Systems 2017, December 4-9, 2017, Long Beach, CA, {USA}}, pages 3391--3401.

\bibitem[Zambaldi et~al., 2019]{zambaldi2018deep}
Zambaldi, V., Raposo, D., Santoro, A., Bapst, V., Li, Y., Babuschkin, I., Tuyls, K., Reichert, D., Lillicrap, T., Lockhart, E., Shanahan, M., Langston, V., Pascanu, R., Botvinick, M., Vinyals, O., and Battaglia, P. (2019).
\newblock Deep reinforcement learning with relational inductive biases.
\newblock In {\em International Conference on Learning Representations}.

\bibitem[Zhao et~al., 2022]{zhao2022compositionalgeneralizationobjectorientedworld}
Zhao, L., Kong, L., Walters, R., and Wong, L. L.~S. (2022).
\newblock Toward compositional generalization in object-oriented world modeling.
\newblock In {\em International Conference on Machine Learning, {ICML} 2022, 17-23 July 2022, Baltimore, Maryland, {USA}}, volume 162 of {\em Proceedings of Machine Learning Research}, pages 26841--26864. {PMLR}.

\bibitem[Zhou et~al., 2022]{zhou2022policyarchitecturescompositionalgeneralization}
Zhou, A., Kumar, V., Finn, C., and Rajeswaran, A. (2022).
\newblock Policy architectures for compositional generalization in control.

\end{thebibliography}
\end{document}